\documentclass[11pt]{article}
\usepackage[final]{acl}
\usepackage{times}
\usepackage{latexsym}
\usepackage[T1]{fontenc}
\usepackage{microtype}
\usepackage{inconsolata}
\usepackage{graphicx}
\usepackage{booktabs}
\usepackage{array}
\usepackage{adjustbox}
\usepackage{amssymb}
\usepackage{xcolor}
\usepackage{colortbl}
\title{Tool Recommendation for Multi-Step Agents:\\ Finding Every Tool a Task Needs}

\author{Sam Partee \quad River Stevens \quad Thierry Damiba \quad Valerie Fanelle \\
  Arcade.dev \\
  \texttt{\{sam, river, thierry, valerie\}@arcade.dev}}

\begin{document}
\maketitle
\begin{abstract}
A multi-step agent task fails when a single tool it needs is missing, yet the tools an agent may need are too many to list: the roughly 8,200 tools on the platform we study have definitions that together exceed the context window of any model we evaluate. We present Tool Recommendation, a deployed system that gives an agent all of them, hundreds of times more than a typical MCP server exposes, through a fixed interface of three tools. It runs at the gateway between agent and tools, executing every call under the end user's own credentials and checking it against the user's consent for each service and the tools the agent may use. We evaluate it on both a public retrieval benchmark and billed end-to-end session cost. On ToolRet, Tool Recommendation places every tool a query needs in the top ten 57\% more often than BM25, a lexical ranking that commercial tool search offers, and compares favorably with published systems up to nine times its size. In paired multi-step workflows, its first recommendation covered every required tool in 93\% of Codex sessions, and against a fixed list of 100 tools it cut input tokens by 17\% with Codex and 40\% with Claude Code and the total bill in all 14 runs, by up to 40\%, with no change in task completion.
\end{abstract}

\section{Introduction}\label{sec:1}

Tool calling has become the standard interface through which large language model (LLM) agents act in the world \citep{schick2023,qin2024,openai2023}. Coding agents work with a handful of tools, such as a file system and a shell, whereas agents that carry out general enterprise work must reach many applications, each exposing many tools. The Model Context Protocol (MCP) made this practical by replacing a custom integration for each application with a single protocol that any MCP client can use \citep{anthropic2024}.

However, an agent can invoke only the tools whose definitions it has been given. Connecting five common MCP servers (GitHub, Slack, Sentry, Grafana and Splunk) loads 58 tool definitions totalling about 55K tokens before the first user turn \citep{anthropic2025a}. In enterprise deployments, a gateway often sits between agents and these servers. It presents their tools behind a single endpoint and handles authorization and governance centrally \citep{saha2026}.

Arcade, the platform this work comes from, is a runtime that executes the tool calls of enterprise AI agents and includes such a gateway. It authorizes each tool call against both the calling user's permissions and the agent's scope. It provides its own catalog of tools as well as the custom tools each organization registers. Organizations give each agent a chosen set of these tools through an Arcade MCP gateway, which works with any MCP client. At the time of our evaluation, Arcade's public catalog contained about 8,200 tools whose definitions total roughly 1.76M tokens, far more than is practical to send on every request.

\begin{figure}[t]
\centering
\includegraphics[width=\columnwidth]{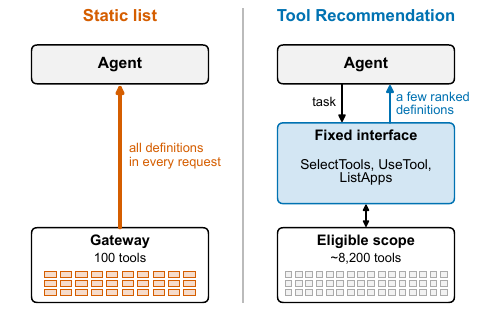}
\setcounter{figure}{0}
\caption{Two ways to give an agent tools: a static list and Tool Recommendation.}
\label{fig:1}
\end{figure}

To let agents work across the applications an enterprise uses, organizations must choose which tools each agent sees, and each common choice gives something up. A curated list keeps requests small but limits the agent to the tasks the list anticipates. The median Arcade gateway is created with 30 tools, under 0.4\% of the full catalog. A longer list widens reach but raises the cost of every request, since each definition is sent again on every turn, and selection accuracy can fall as the list grows \citep{anthropic2025b,kate2025}. Fine-tuning a model on its tools teaches it when and how to call them \citep{patil2024,qin2024,wang2025}, but it assumes a fixed tool set. Enterprise tool sets are not fixed. Providers add and change tools, and each organization registers its own, so a curated list or a fine-tuned model falls behind the catalog it was built from. Arcade's catalog, for example, can change daily. An agent therefore needs the whole catalog, at lower cost than a static list, without training on each organization's tools.

Several deployed systems now take a different approach, in which the agent searches the catalog and calls what it finds (Figure~\ref{fig:1}). Microsoft's Azure AI Foundry ships it as a pair of meta-tools \citep{balaji2026}, PayPal runs it in production across more than 2,000 tools \citep{saha2026}, and two research systems for MCP servers, RAG-MCP and MCP-Zero, publish variants \citep{gan2025,fei2025}. Anthropic and OpenAI offer deferred loading, which keeps definitions out of the model's context even though the client still sends every definition on every request \citep{anthropic2025b,openai2026}. Search gives an agent the catalog without training on it, but the agent can then do only what the search returns. Its quality bounds the agent, and retrieval errors already account for nearly half of all failures in a live-server benchmark of 527 tools \citep{mo2025}.

The published evaluations of these systems measure one side of the design at a time. Vendor reports count the tokens a system avoids sending \citep{anthropic2025a,balaji2026,saha2026}, but token reductions predict billed cost poorly, in large part because prompt caching dominates the bill \citep{weinberger2026}. Recommendation quality is reported less often: PayPal evaluates on a 49-query internal benchmark, and Microsoft publishes Recall@10 by category on ToolRet, a public benchmark \citep{shi2025}. We report both for a single deployed system.

In this work, we present Tool Recommendation, Arcade's capability for giving agents the full catalog through a small, fixed interface. The agent receives three tools instead of a tool list. It describes a task to SelectTools and receives the definitions of the tools that task needs, then runs them through UseTool under the calling user's own authorization. ListApps tells the agent which applications it can reach. The system recommends from Arcade's public catalog and the tools an organization registers. We frame the problem as recommendation rather than retrieval alone: the system matches tasks to tools as a recommender matches users to items, over a catalog that changes daily.

We evaluate whether it finds every tool a task needs, and what it costs. On ToolRet (7,961 queries over 44,453 tools), it places every needed tool in the top ten for 45.2\% of queries, against 28.8\% for BM25 \citep{robertson2009}, and in end-to-end Codex sessions its first recommendation covered every tool the workflow required in 304 of 327 sessions. In paired, end-to-end runs of multi-step enterprise workflows against a 100-tool static list, it lowered input tokens by 17\% with Codex and 40\% with Claude Code, and the total bill by 13.6\% across both clients, with every run cheaper and no change in task completion.

Our contributions are as follows. (i) We describe Tool Recommendation, a deployed system that serves an 8,200-tool catalog through three tools, without fine-tuning on customer tools. It runs at the gateway rather than in the client, where it executes every call under the end user's own credentials and authorizes it per service and per tool (\S\ref{sec:2}). (ii) We evaluate whether it finds every tool a task needs, with Completeness@10 and Recall@10 on ToolRet, as lift over BM25 and against published systems, and in end-to-end sessions (\S\ref{sec:4.1}). (iii) We measure session cost on two agent clients in paired, end-to-end evaluations priced at list rates, and characterize when recommendation lowers cost and when a static list remains cheaper (\S\ref{sec:4.2}, \S\ref{sec:4.3}).

\section{System}\label{sec:2}

Tool Recommendation sits between the agent and Arcade's tool catalog. The agent never receives the catalog. It receives a small, fixed set of tools through which it requests, runs and reads the tools a task requires (Figure~\ref{fig:2}). Recommendation runs inside Arcade's execution runtime, so every tool it recommends stays within the gateway's allowlist and access policy and runs under the calling user's own authorization, with credentials kept out of the agent's context (\S\ref{sec:2.2}, \S\ref{sec:2.4}). An enterprise can therefore expose its full catalog without widening what any agent is allowed to do.

\begin{figure*}[t]
\centering
\includegraphics[width=\textwidth]{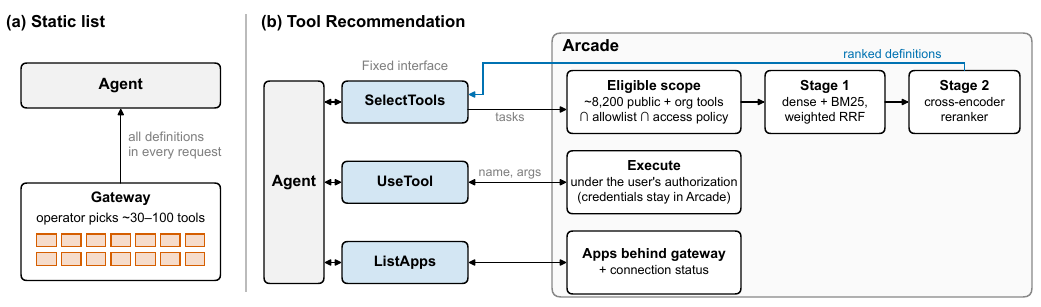}
\setcounter{figure}{1}
\caption{System overview.}
\label{fig:2}
\end{figure*}

\subsection{Interface}\label{sec:2.1}

\paragraph{Terminology.}
A static list is a gateway that lists a fixed set of tools and sends every definition on every request. A discovery-enabled gateway instead exposes the interface below and adds tools as the agent requests them.

Tool Recommendation gives an agent three tools over the Model Context Protocol, so any MCP client can use them. SelectTools takes one or more task descriptions in natural language and returns, for each, the five highest-ranked tool definitions with full input schemas. UseTool runs a returned tool with the given arguments. ListApps reports which applications sit behind the gateway and whether the user has connected each one. With the gateway's own authorization tool, a client lists four tools whose definitions total 993 tokens, about 1,800 times fewer than the catalog's.

\subsection{Catalog and Scope}\label{sec:2.2}

Recommendation operates over Arcade's public catalog, 8,252 tools as of 2026-09-14, together with any tools an organization registers. The catalog can change daily, so we round it to about 8,200 elsewhere. Each request is restricted to an eligible scope: the tools on the calling gateway's allowlist, filtered further by any access policy that applies in the calling context. Recommendation ranks only within this scope, so tools outside the allowlist are excluded before ranking.

\subsection{Recommendation}\label{sec:2.3}

In a first stage, Recommendation retrieves candidate tools for each task description from the eligible scope by fusing lexical (BM25) and dense retrieval. A second stage reorders the leading candidates with a cross-encoder that reads task and tool together, whenever the top candidates score close together, which in production is more than 99.9\% of requests. Both models are fine-tuned by Arcade on examples of tasks and the tools that accomplish them. In production, the ranking service answers a request in 131 ms at the median and 292 ms at the 95th percentile, of which reranking takes 62 ms and 195 ms, and a SelectTools call takes about 1.4 s at the median end to end.

\subsection{Execution}\label{sec:2.4}

UseTool executes each tool under the end user's own authorization. Provider credentials are held by Arcade and never enter the agent's context. Authorization is checked twice for every call: the user must have granted consent for the tool's service, and the tool must lie within the gateway's eligible scope (\S\ref{sec:2.2}). When a tool requires consent the user has not yet granted, the model receives an authorization link in place of a result, and the call can be repeated once the user has approved it.

\section{Methodology}\label{sec:3}

Table~\ref{tab:1} lists the evaluations and the section that reports each. Every number in \S\ref{sec:4} comes from one of them.

\begin{table*}[t]
\centering
\small
\setlength{\tabcolsep}{4pt}
\begin{adjustbox}{max width=\textwidth}
\begin{tabular}{l>{\raggedright\arraybackslash}p{0.317\textwidth}>{\raggedright\arraybackslash}p{0.317\textwidth}l}
\toprule
Evaluation & Measures & Setup & Section \\
\midrule
Ranking quality & Whether the ranking returns every tool a query needs & ToolRet, 7,961 queries over 44,453 tools, against BM25 in the same harness & \S\ref{sec:4.1} \\
Session cost, Codex & Input tokens, total bill and completion against a 100-tool static list & 30 multi-step workflows, 11 runs, gpt-5.6-luna & \S\ref{sec:4.2} \\
Session cost, Claude Code & Same, on a second client & 34 workflows, 3 runs, Sonnet 5 & \S\ref{sec:4.2} \\
Short lookups & Input tokens on single-tool tasks against small catalogs & 14 tasks, 5 runs each, catalogs of 8--37 tools, Codex & \S\ref{sec:4.2} \\
Cost model & Input cost of a typical session & Nine models at list prices, production search rate, large results kept out of context & \S\ref{sec:4.3} \\
\bottomrule
\end{tabular}
\end{adjustbox}
\setcounter{table}{0}
\caption{The evaluations and where each result is reported.}
\label{tab:1}
\end{table*}

\subsection{Recommendation Quality}\label{sec:3.1}

We use ToolRet \citep{shi2025}, a public benchmark of 7,961 queries over 44,453 tools in three categories: web APIs, code functions and customized applications. We report Recall@10, the metric published for a deployed system \citep{balaji2026}, and Completeness@10, the fraction of queries with every relevant tool in the top ten \citep{shi2025}. Completeness matters because a task missing one tool in a chain fails. An agent asked to find a Linear issue and then edit it fails without the find tool, yet Recall@10 scores that result at 50\%. In Arcade's own optimized toolkits, 504 of 1,181 tools need an identifier that only a prior call produces. Each query searches the full index with the fixed instruction production sends.

Absolute scores do not transfer between harnesses (Appendix~\ref{app:B}), so we report each result as lift over a BM25 baseline run in ours. We evaluate Tool Recommendation with the fusion weight production uses (\ensuremath{\theta} = 0.5), indexing each tool's name, description and argument documentation.

\subsection{Session Cost}\label{sec:3.2}

We measure cost end to end on multi-step enterprise workflows in a seeded test tenant spanning Gmail, Slack, Google Drive, Salesforce, Jira, Confluence and Zendesk, with synthetic records. A workflow passes when checks confirm that the agent read the right sources and made the required writes. Each workflow runs in two conditions: a static gateway that lists 100 tools, and a discovery-enabled gateway that exposes the interface of \S\ref{sec:2.1} over the same 100 tools. A static list of the full catalog does not fit in the context window of any model we evaluate, so 100 tools is the comparison.

With Codex 0.144.6 and gpt-5.6-luna, we ran the evaluation's 30 workflows eleven times, one attempt per workflow per run, in four variants (Appendix~\ref{app:A}), with a proxy recording every exchange. In eight of the eleven runs, as in the evaluation's original design, the gateway's instructions to the agent reached only the recommendation gateway; the runs with instructions on both gateways gave similar results (Table~\ref{tab:4}). With Claude Code 2.1.291 and Sonnet 5, its own tool search off, we ran 34 workflows: the 21 of the 30 that use neither Jira nor Confluence, twice each, and 13 workflows that return large or mixed-size tool results, three times each. In the Claude Code runs the recommendation gateway also kept tool results over 14 KB out of the conversation, passing the model a digest it can read further on request. Separately, 14 short lookup tasks, most needing one tool, ran five times each against catalogs of 8--37 tools.

\subsection{Modeled Cost}\label{sec:3.3}

To estimate cost beyond the measured workflows, we use a token model of a typical session: three user messages, each answered with eight tool calls, a search rate of 0.12 searches per tool call over 30 days of production use during the system's free research preview, tool-result sizes from 1,465 production tool calls, and a static list of 100 tools at the catalog's average definition size. It counts input tokens and applies each vendor's caching rules and list prices for nine models. It assumes the client keeps large tool results out of the conversation and passes the model a short digest in their place, so that both arms receive 2,909 tokens per tool call, the average over those 1,465 calls including the parts agents read back, instead of 14,319.

\subsection{Statistical Procedure}\label{sec:3.4}

We compare paired sessions, in which one workflow and attempt runs in both conditions. For session cost we report the change in total input tokens and in the total bill over all pairs, which is what an organization running these workflows would pay. Dollar cost uses list prices: gpt-5.6-luna at \$0.20, \$0.02 and \$1.20 per million uncached input, cached input and output tokens, and Claude Code's own reported cost. Because runs repeat the same workflows, 95\% intervals come from 10,000 bootstrap resamples of whole workflows, and p-values from permutation tests that swap the two conditions for all of a workflow's pairs. Breakdowns by search count and by run use the median change per session over pairs that passed in both conditions. Completion is compared with an exact McNemar test.

\section{Results}\label{sec:4}

Tool Recommendation finds more of the tools a task needs than BM25, and on multi-step work it sends less input and costs less than a 100-tool static list without lowering completion. On ToolRet it places every tool a query needs in the top ten for 45.2\% of queries, against 28.8\% for BM25, and in end-to-end sessions its first recommendation covered every tool the workflow required in 93\% of cases (\S\ref{sec:4.1}). Across two clients it cut input tokens by 17\% to 40\% and the total bill by 13.6\%, and every run was cheaper (\S\ref{sec:4.2}). A static list stays cheaper for small catalogs, short lookups and search-heavy sessions (\S\ref{sec:4.2}, \S\ref{sec:4.3}).

\subsection{Tool Recommendation completes more tool chains than BM25}\label{sec:4.1}

On Completeness@10, the stricter test that every tool a query needs appears in the top ten, Tool Recommendation reaches 45.2 overall against 28.8 for BM25, a lexical ranking that commercial tool search offers \citep{anthropic2025b}, in the same harness, a 57\% relative gain (Table~\ref{tab:2}), and it leads BM25 on both metrics in every category, by 79\% on code Completeness@10 (Table~\ref{tab:5}). Its overall Recall@10, counting every query equally, is 1.49 times BM25's (55.2 against 37.0; 95\% CI 1.35 to 1.78). Against the published systems, its absolute scores are the highest on web and code tools and on average, on both metrics, while three published baselines with 6 to 9 times as many parameters (7.2B to 10.4B against Tool Recommendation's 1.2B) score higher on customized applications. Measured as lift over the BM25 of each system's own harness, the comparison \S\ref{sec:3.1} recommends, it leads on code and customized applications and on average, and bge-reranker-v2-gemma and Foundry lead on web.

\begin{table*}[t]
\centering
\small
\setlength{\tabcolsep}{4pt}
\begin{adjustbox}{max width=\textwidth}
\begin{tabular}{llrrrrrrr}
\toprule
 &  &  & \multicolumn{4}{c}{Recall@10} & \multicolumn{2}{c}{Completeness@10} \\
\cmidrule(lr){4-7} \cmidrule(lr){8-9}
System & Harness & Parameters & Web & Code & Customized & Average & Average & Overall \\
\midrule
Tool Recommendation & Arcade & 1.2B & \textbf{56.8} & \textbf{57.3} & 42.7 & \textbf{52.2} & \textbf{42.4} & \textbf{45.2} \\
\emph{lift over own BM25} &  &  & +40.7\% & \textbf{+74.6\%} & \textbf{+60.7\%} & \textbf{+57.2\%} & \textbf{+66.7\%} & \textbf{+57.0\%} \\
bge-reranker-v2-gemma & ToolRet\textsuperscript{\dag} & 10.4B\textsuperscript{\ddag} & 45.9 & 38.2 & \textbf{49.4} & 44.5 & 34.1 & -- \\
\emph{lift over own BM25} &  &  & +86.6\% & +35.4\% & +52.6\% & +56.7\% & +53.9\% &  \\
Foundry tool search & Microsoft\textsuperscript{\dag} & -- & 46.0 & 39.6 & 41.4 & 42.3 & -- & -- \\
\emph{lift over own BM25} &  &  & \textbf{+86.8\%} & +40.1\% & +27.7\% & +48.9\% &  &  \\
NV-Embed-v1 & ToolRet\textsuperscript{\dag} & 7.9B & 39.2 & 40.5 & 45.9 & 41.8 & 32.1 & -- \\
\emph{lift over own BM25} &  &  & +59.0\% & +43.3\% & +41.8\% & +47.3\% & +44.8\% &  \\
GritLM-7B & ToolRet\textsuperscript{\dag} & 7.2B & 34.3 & 30.4 & 49.3 & 38.0 & 29.4 & -- \\
\emph{lift over own BM25} &  &  & +39.2\% & +7.7\% & +52.3\% & +33.8\% & +32.7\% &  \\
BM25 & Arcade & -- & 40.4 & 32.8 & 26.5 & 33.2 & 25.4 & 28.8 \\
BM25s & ToolRet\textsuperscript{\dag} & -- & 24.6 & 28.2 & 32.4 & 28.4 & 22.2 & -- \\
\bottomrule
\end{tabular}
\end{adjustbox}
\setcounter{table}{1}
\caption{ToolRet results. Completeness@10 is the share of queries with every needed tool in the top ten. Each system's lift is over the BM25 run in its own harness, shown below it. Parameters count every model a system runs; Foundry's is a sparse pipeline with no reported model. Average is the mean of the three categories; Overall weights each query equally. Bold marks the best absolute score and the best lift in each column. \textsuperscript{\dag}Quoted from \citet[Table~4, without instructions]{shi2025} and \citet{balaji2026}. Scores from different harnesses are not directly comparable (\S\ref{sec:3.1}); lift over each harness's own BM25 is. \textsuperscript{\ddag}bge-reranker-v2-gemma (2.5B) reranks NV-Embed-v1's (7.9B) results. Table~\ref{tab:5} gives every category on both metrics.}
\label{tab:2}
\end{table*}

\paragraph{Most misses come from how the agent splits the task.}
A recommendation can fail in two ways. The system can return the wrong tools for a task the agent asked about, or the agent can fail to ask for a tool it needs. Over the eleven Codex runs, the first SelectTools call returned every tool the workflow required in 304 of 327 sessions. Of the 24 required calls it missed, 13 were never returned by any search and 10 came back later for a task the first call never asked about. Only one was a recommendation error. The larger source of misses is therefore how the agent decomposes the task, not the ranking.

\subsection{Recommendation lowers input and cost on multi-step work}\label{sec:4.2}

Across both clients, recommendation lowered the total bill in every one of the 14 runs, by 2.6\% to 40.0\% per run and by 13.6\% overall (95\% CI 7.4\% to 18.4\%, p = 0.006; Table~\ref{tab:3}). It lowered total input tokens by 17.2\% with Codex and 39.6\% with Claude Code, and the median Codex session used 19.0\% less input. Completion did not differ: 276 of 327 Codex sessions passed with recommendation and 273 with the static list (exact McNemar p = 0.80), and 74 against 76 of 81 Claude Code sessions. By client, the bill fell 5.5\% with Codex, an interval that on its own includes zero, and 21.8\% with Claude Code, reaching 40.0\% on the workflows with large tool results. In most Claude Code sessions, the tools found by an earlier session were still listed, so the agent seldom had to search.

\begin{table*}[t]
\centering
\small
\setlength{\tabcolsep}{4pt}
\begin{adjustbox}{max width=\textwidth}
\begin{tabular}{>{\raggedright\arraybackslash}p{0.135\textwidth}rrrrr}
\toprule
Client (model) & Pairs (workflows) & Input tokens [95\% CI] & Total bill [95\% CI] & Runs cheaper & Passed, rec / static \\
\midrule
Codex (gpt-5.6-luna) & 327 (30) & \ensuremath{-}17.2\% [\ensuremath{-}25.3, \ensuremath{-}8.6] & \ensuremath{-}5.5\% [\ensuremath{-}12.0, +1.3] & 11 of 11 & 276 / 273 of 327 \\
Claude Code (Sonnet 5) & 81 (34) & \ensuremath{-}39.6\% [\ensuremath{-}49.3, \ensuremath{-}24.5] & \ensuremath{-}21.8\% [\ensuremath{-}28.9, \ensuremath{-}10.9] & 3 of 3 & 74 / 76 of 81 \\
Both clients & 408 (43 distinct) & -- & \ensuremath{-}13.6\% [\ensuremath{-}18.4, \ensuremath{-}7.4] & 14 of 14 & -- \\
\bottomrule
\end{tabular}
\end{adjustbox}
\setcounter{table}{2}
\caption{Change in total input tokens and total bill over all paired sessions, recommendation against the 100-tool static list. Intervals resample whole workflows. The clients share 21 workflows, so the 30 and the 34 are 43 distinct; the pooled row weights the two clients equally.}
\label{tab:3}
\end{table*}

The saving depends on how often the agent searches (Figure~\ref{fig:3}). Codex workflows that searched once changed input by a median \ensuremath{-}28.2\% (95\% CI \ensuremath{-}34.2\% to \ensuremath{-}20.0\%, n = 153), while those that searched three times or more used more input than the static list (+32.3\% for three, +43.8\% for four or more). Likewise, on the 14 short lookups against catalogs of 8--37 tools, recommendation increased input by a median 22.0\% (95\% CI 13.4\% to 30.4\%): each search costs a turn that a short list does not need.

\begin{figure}[t]
\centering
\includegraphics[width=\columnwidth]{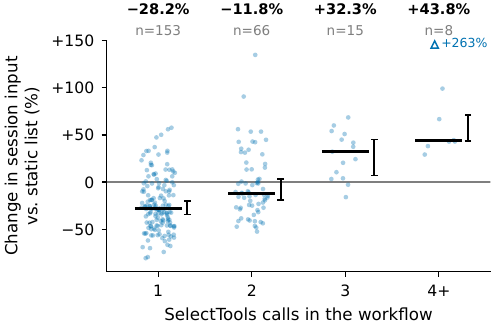}
\setcounter{figure}{2}
\caption{Change in session input by number of SelectTools calls per workflow. Bars: 95\% intervals resampling whole workflows.}
\label{fig:3}
\end{figure}

\subsection{Cost at production scale}\label{sec:4.3}

For a typical session priced across nine models, the cost model predicts an input-cost saving of 18\% to 21\% against a 100-tool static list, rising to 37\% for sessions of a single message. Keeping results out of context does not change the dollars recommendation saves, since both arms carry the same results, but it shrinks the bill around them, so the percentage saved is about three times what it is when every result stays in the conversation. A static list becomes the cheaper choice only below about 25 to 28 tools, or when the agent searches more than about 0.63 times per tool call, five times the observed rate.

\section{Discussion}\label{sec:5}

\paragraph{Savings and reach together.}
Tool Recommendation gives an agent the whole eligible catalog and, on multi-step work, still costs less than a 100-tool static list. What decides the saving is how often the agent searches: each search adds a turn that re-sends the conversation, so a single search pays for itself and three do not (Figure~\ref{fig:3}). A static list remains the better choice for small, stable tool sets and short tasks. Recommendation pays as the catalog grows and when an agent can find what it needs in one search.

\paragraph{Input tokens overstate the bill.}
Providers cache the request prefix and bill cached tokens at about a tenth of the full rate, while each tool result arrives uncached at the end of the request. With Codex, 85\% of input with recommendation and 87\% with the static list was served from the prompt cache. Most of the input that recommendation removes is therefore cached, which is why a 17\% to 40\% cut in input tokens became a 5\% to 22\% cut in the bill (\S\ref{sec:4.2}). Claims about cost should therefore be made in dollars, with the client's caching behavior stated.

\paragraph{Completeness matters more than recall.}
An agent missing one step of a chain fails the task. Accordingly, Tool Recommendation's lift over BM25 was larger in Completeness@10 than in Recall@10 in every category (Table~\ref{tab:5}), and largest on code in absolute points. This suggests judging a ranking by whether it returns the complete set of tools a task needs.

\section{Related Work}\label{sec:6}

\paragraph{Tool retrieval.}
COLT optimizes for retrieving the complete set of tools a task requires \citep{qu2024}, the reasoning behind our use of Completeness@10. Other work improves query and tool representations \citep{chen2024,moon2024,lumer2024}, scales retrieval to thousands of MCP servers \citep{lumer2025}, aligns queries with tools through preference optimization \citep{zhang2024}, pre-selects tools with a small classifier for an on-device agent \citep{erdogan2024}, folds retrieval into generation \citep{wang2025}, studies reranking \citep{zheng2024}, or models relationships between tools \citep{du2024,hong2026}. Benchmarks of tool use measure planning, retrieval and calling together \citep{li2023,shen2024}; on Seal-Tools, a fine-tuned agent's tool recall falls from 92\% on single-tool instances to 77\% on multi-tool ones \citep{wu2024}. We adopt the two-stage structure of large-scale recommenders \citep{covington2016} over a catalog whose items must be recommendable on the day they appear. Most models degrade as the tool set grows \citep{lei2025}, and Recall@K at large K can carry no selectivity over random selection \citep{repantis2026}, so we report k = 10.

\paragraph{Deployed systems and cost.}
Microsoft's Azure AI Foundry exposes tool\_search and call\_tool over a sparse retrieval pipeline and publishes Recall@10 on ToolRet \citep{balaji2026}. Anthropic's tool search and deferred loading reduce definition tokens by more than 85\% \citep{anthropic2025a,anthropic2025b}. OpenAI and AWS ship equivalents \citep{openai2026,aws2026}, but we found no published evaluation for either. \citet{saha2026} describe SCOUT at PayPal, which fuses BM25 and dense retrieval over more than 2,000 tools and filters tools by user authorization. On 45 evaluable queries of a 49-query internal benchmark, SCOUT reports a Hit@5 of 95.6\%.

\citet{li2026} describe a production tool gateway serving more than 3,000 tools, reporting 98\% top-15 recall on an internal evaluation and 23.8 times fewer tokens than a baseline that mounts 372 tools. However, whether removing tools from context helps is itself contested. Specifically, models often attend to the right tool and still choose wrongly \citep{chen2026}, progressive disclosure buys context rather than capability \citep{he2026}, schema compression helps only under tight context budgets \citep{sakizli2026}, and reductions in tool-output tokens correlate weakly with billed cost \citep{weinberger2026}. This work reports both recommendation quality on a public benchmark and billed session cost for one deployed system.

\section{Conclusion}\label{sec:7}

We presented Tool Recommendation, a deployed system that gives an agent every tool on an 8,200-tool platform, hundreds of times more than a typical MCP server exposes, through a fixed interface of three tools. It runs at the gateway between agent and tools, executing every call under the end user's own credentials and consent. We measured both how completely it finds the tools a task needs and what it costs. On ToolRet, it places every tool a query needs in the top ten 57\% more often than BM25 and compares favorably with published systems up to nine times its size. In end-to-end Codex sessions, its first recommendation covered every required tool 93\% of the time. Against a fixed list of 100 tools on multi-step workflows, it cut input tokens by 17\% with Codex and 40\% with Claude Code, and the total bill in every one of 14 runs, by up to 40\% and by 13.6\% across both clients, with no change in task completion. For clients that keep large tool results out of context, a cost model calibrated on production traffic predicts about 20\% lower input cost for a typical session. A static list remains cheaper only for small catalogs, short lookups and search-heavy sessions. Giving an agent every tool it may need, rather than a curated hundred, therefore costs less, not more.

\section*{Limitations}\label{sec:limitations}

\paragraph{Tool Recommendation is only available through Arcade.}
It works only with tools served by an Arcade gateway, so others cannot reproduce the session results outside Arcade.

\paragraph{ToolRet measures ranking, not agent behavior.}
Its static prose documentation cannot express multi-turn context, authorization state or tools that depend on an earlier call's output. Published baselines were run in other harnesses, so their absolute scores are not directly comparable with ours (\S\ref{sec:3.1}). We compare lift over each harness's own BM25.

\paragraph{The session-cost evaluation is limited in scope.}
It covers two clients on 30 to 34 seeded workflows, against a 100-tool static list, at list prices.

\textbf{The cost model rests on assumed session shapes} and on search rates and result sizes from limited production traffic. It assumes clients keep large tool results out of context, does not price the extra turns an agent spends reading a stored result back, and counts input tokens only.

\paragraph{Cost depends on the client.}
Clients differ in how they load and re-read a tool list that grows during use, which changes what each search costs.

\section*{Acknowledgments}\label{sec:acknowledgments}

The authors used a text-to-text generative AI tool to draft and revise the text of this paper. The authors verified all content and take responsibility for it.

\bibliography{refs}

\begin{thebibliography}{40}
\providecommand{\natexlab}[1]{#1}

\bibitem[{{Anthropic}(2024)}]{anthropic2024}
{Anthropic}. 2024.
\newblock \href {https://www.anthropic.com/news/model-context-protocol}
  {{Introducing the Model Context Protocol}}.
\newblock Anthropic news, November 25, 2024.

\bibitem[{{Anthropic}(2025{\natexlab{a}})}]{anthropic2025a}
{Anthropic}. 2025{\natexlab{a}}.
\newblock \href {https://www.anthropic.com/engineering/advanced-tool-use}
  {{Introducing advanced tool use on the Claude Developer Platform}}.
\newblock Anthropic Engineering blog, November 24, 2025.

\bibitem[{{Anthropic}(2025{\natexlab{b}})}]{anthropic2025b}
{Anthropic}. 2025{\natexlab{b}}.
\newblock \href
  {https://platform.claude.com/docs/en/agents-and-tools/tool-use/tool-search-tool}
  {{Tool search tool}}.
\newblock Claude Developer Platform documentation. Accessed 2026-10-06.

\bibitem[{{AWS}(2026)}]{aws2026}
{AWS}. 2026.
\newblock \href
  {https://docs.aws.amazon.com/bedrock-agentcore/latest/devguide/gateway-using-mcp-semantic-search.html}
  {{Search for tools in your AgentCore gateway with a natural language query}}.
\newblock Amazon Bedrock AgentCore Developer Guide. Accessed 2026-10-12.

\bibitem[{Balaji and Sathe(2026)}]{balaji2026}
V.~Balaji and S.~Sathe. 2026.
\newblock \href
  {https://commandline.microsoft.com/tool-search-toolboxes-foundry/} {{Tool
  search: Finding the right tool at the right time}}.
\newblock Command Line (Microsoft), July 29, 2026.

\bibitem[{Chen(2026)}]{chen2026}
Shiyang Chen. 2026.
\newblock \href {https://arxiv.org/abs/2606.16364} {{Looking is not picking: An
  attention-segment account of tool-selection failures in LLM agents}}.
\newblock \emph{Preprint}, arXiv:2606.16364.

\bibitem[{Chen et~al.(2024)Chen, Yoon, Sachan, Wang, Cohen-Addad, Bateni, Lee,
  and Pfister}]{chen2024}
Yanfei Chen, Jinsung Yoon, Devendra~Singh Sachan, Qingze Wang, Vincent
  Cohen-Addad, Mohammadhossein Bateni, Chen-Yu Lee, and Tomas Pfister. 2024.
\newblock \href {https://arxiv.org/abs/2408.01875} {{Re-Invoke: Tool invocation
  rewriting for zero-shot tool retrieval}}.
\newblock In \emph{Findings of the Association for Computational Linguistics:
  EMNLP 2024}.

\bibitem[{Covington et~al.(2016)Covington, Adams, and Sargin}]{covington2016}
Paul Covington, Jay Adams, and Emre Sargin. 2016.
\newblock {Deep neural networks for YouTube recommendations}.
\newblock In \emph{Proceedings of the 10th ACM Conference on Recommender
  Systems}, pages 191--198.

\bibitem[{Du et~al.(2024)Du, Wei, and Zhang}]{du2024}
Yu~Du, Fangyun Wei, and Hongyang Zhang. 2024.
\newblock \href {https://arxiv.org/abs/2402.04253} {{AnyTool: Self-reflective,
  hierarchical agents for large-scale API calls}}.
\newblock In \emph{Proceedings of the 41st International Conference on Machine
  Learning}.

\bibitem[{Erdogan et~al.(2024)Erdogan, Lee, Jha, Kim, Tabrizi, Moon, Hooper,
  Anumanchipalli, Keutzer, and Gholami}]{erdogan2024}
Lutfi~Eren Erdogan, Nicholas Lee, Siddharth Jha, Sehoon Kim, Ryan Tabrizi,
  Suhong Moon, Coleman Hooper, Gopala Anumanchipalli, Kurt Keutzer, and Amir
  Gholami. 2024.
\newblock \href {https://arxiv.org/abs/2409.00608} {{TinyAgent: Function
  calling at the edge}}.
\newblock In \emph{Proceedings of the 2024 Conference on Empirical Methods in
  Natural Language Processing: System Demonstrations}.

\bibitem[{Fei et~al.(2025)Fei, Zheng, and Feng}]{fei2025}
Xiang Fei, Xiawu Zheng, and Hao Feng. 2025.
\newblock \href {https://arxiv.org/abs/2506.01056} {{MCP-Zero: Active tool
  discovery for autonomous LLM agents}}.
\newblock \emph{Preprint}, arXiv:2506.01056.

\bibitem[{Gan and Sun(2025)}]{gan2025}
Tiantian Gan and Qiyao Sun. 2025.
\newblock \href {https://arxiv.org/abs/2505.03275} {{RAG-MCP: Mitigating prompt
  bloat in LLM tool selection via retrieval-augmented generation}}.
\newblock \emph{Preprint}, arXiv:2505.03275.

\bibitem[{He et~al.(2026)He, Zhao, Wang, and Chen}]{he2026}
Yifeng He, Yinzhe Zhao, Jicheng Wang, and Hao Chen. 2026.
\newblock \href {https://arxiv.org/abs/2607.17598} {{Is progressive disclosure
  all you need for long-context agents?}}
\newblock \emph{Preprint}, arXiv:2607.17598.

\bibitem[{Hong et~al.(2026)Hong, Dong, Yu, and Jiang}]{hong2026}
Xinyi Hong, Pinjun Dong, Xinyang Yu, and Binyan Jiang. 2026.
\newblock \href {https://arxiv.org/abs/2607.25718} {{Tools are not islands:
  Set-level tool retrieval for LLM agents via query-conditioned hyperedge
  prediction}}.
\newblock \emph{Preprint}, arXiv:2607.25718.

\bibitem[{Kate et~al.(2025)Kate, Pedapati, Basu, Rizk et~al.}]{kate2025}
Kiran Kate, Tejaswini Pedapati, Kinjal Basu, Yara Rizk, et~al. 2025.
\newblock \href {https://arxiv.org/abs/2505.10570} {{LongFuncEval: Measuring
  the effectiveness of long context models for function calling}}.
\newblock \emph{Preprint}, arXiv:2505.10570.

\bibitem[{Lei et~al.(2025)Lei, Yang, Sun, and Lin}]{lei2025}
Fei Lei, Yibo Yang, Wenxiu Sun, and Dahua Lin. 2025.
\newblock \href {https://arxiv.org/abs/2508.16260} {{MCPVerse: An expansive,
  real-world benchmark for agentic tool use}}.
\newblock \emph{Preprint}, arXiv:2508.16260.

\bibitem[{Li et~al.(2023)Li, Zhao, Yu, Song, Li, Yu, Li, Huang, and
  Li}]{li2023}
Minghao Li, Yingxiu Zhao, Bowen Yu, Feifan Song, Hangyu Li, Haiyang Yu, Zhoujun
  Li, Fei Huang, and Yongbin Li. 2023.
\newblock \href {https://arxiv.org/abs/2304.08244} {{API-Bank: A comprehensive
  benchmark for tool-augmented LLMs}}.
\newblock In \emph{Proceedings of the 2023 Conference on Empirical Methods in
  Natural Language Processing}.

\bibitem[{Li et~al.(2026)Li, Song, Zuo et~al.}]{li2026}
Mingxin Li, Enge Song, Yueshang Zuo, et~al. 2026.
\newblock \href {https://arxiv.org/abs/2607.15593} {{Scalable LLM agent tool
  access in the cloud}}.
\newblock \emph{Preprint}, arXiv:2607.15593.

\bibitem[{Lumer et~al.(2025)Lumer, Gulati, Subbiah, Basavaraju, and
  Burke}]{lumer2025}
Elias Lumer, Anmol Gulati, Vamse~Kumar Subbiah, Pradeep~Honaganahalli
  Basavaraju, and James~A. Burke. 2025.
\newblock \href {https://arxiv.org/abs/2505.06416} {{ScaleMCP: Dynamic and
  auto-synchronizing Model Context Protocol tools for LLM agents}}.
\newblock \emph{Preprint}, arXiv:2505.06416.

\bibitem[{Lumer et~al.(2024)Lumer, Subbiah, Burke, Basavaraju, and
  Huber}]{lumer2024}
Elias Lumer, Vamse~Kumar Subbiah, James~A. Burke, Pradeep~Honaganahalli
  Basavaraju, and Austin Huber. 2024.
\newblock \href {https://arxiv.org/abs/2410.14594} {{Toolshed: Scale
  tool-equipped agents with advanced RAG-tool fusion and tool knowledge
  bases}}.
\newblock \emph{Preprint}, arXiv:2410.14594.

\bibitem[{Mo et~al.(2025)Mo, Zhong, Chen, Yuan, Chen, Lu, Lin, He, Han, and
  Sun}]{mo2025}
Guozhao Mo, Wenliang Zhong, Jiawei Chen, Qianhao Yuan, Xuanang Chen, Yaojie Lu,
  Hongyu Lin, Ben He, Xianpei Han, and Le~Sun. 2025.
\newblock \href {https://arxiv.org/abs/2508.01780} {{LiveMCPBench: Can agents
  navigate an ocean of MCP tools?}}
\newblock \emph{Preprint}, arXiv:2508.01780.

\bibitem[{Moon et~al.(2024)Moon, Jha, Erdogan, Kim, Lim, Keutzer, and
  Gholami}]{moon2024}
Suhong Moon, Siddharth Jha, Lutfi~Eren Erdogan, Sehoon Kim, Woosang Lim, Kurt
  Keutzer, and Amir Gholami. 2024.
\newblock \href {https://arxiv.org/abs/2409.02141} {{Efficient and scalable
  estimation of tool representations in vector space}}.
\newblock \emph{Preprint}, arXiv:2409.02141.

\bibitem[{{OpenAI}(2023)}]{openai2023}
{OpenAI}. 2023.
\newblock \href
  {https://openai.com/index/function-calling-and-other-api-updates/} {{Function
  calling and other API updates}}.
\newblock OpenAI blog, June 13, 2023.

\bibitem[{{OpenAI}(2026)}]{openai2026}
{OpenAI}. 2026.
\newblock \href
  {https://developers.openai.com/api/docs/guides/tools-tool-search} {{Tool
  search}}.
\newblock OpenAI API documentation. Accessed 2026-10-06.

\bibitem[{Patil et~al.(2024)Patil, Zhang, Wang, and Gonzalez}]{patil2024}
Shishir~G. Patil, Tianjun Zhang, Xin Wang, and Joseph~E. Gonzalez. 2024.
\newblock \href {https://arxiv.org/abs/2305.15334} {{Gorilla: Large language
  model connected with massive APIs}}.
\newblock In \emph{Advances in Neural Information Processing Systems 37}.

\bibitem[{Qin et~al.(2024)Qin, Liang, Ye, Zhu, Yan, Lu, Lin, Cong, Tang, Qian,
  Zhao, Hong, Tian, Xie, Zhou, Gerstein, Li, Liu, and Sun}]{qin2024}
Yujia Qin, Shihao Liang, Yining Ye, Kunlun Zhu, Lan Yan, Yaxi Lu, Yankai Lin,
  Xin Cong, Xiangru Tang, Bill Qian, Sihan Zhao, Lauren Hong, Runchu Tian,
  Ruobing Xie, Jie Zhou, Mark Gerstein, Dahai Li, Zhiyuan Liu, and Maosong Sun.
  2024.
\newblock \href {https://arxiv.org/abs/2307.16789} {{ToolLLM: Facilitating
  large language models to master 16000+ real-world APIs}}.
\newblock In \emph{The Twelfth International Conference on Learning
  Representations}.

\bibitem[{Qu et~al.(2024)Qu, Dai, Wei, Cai, Wang, Yin, Xu, and Wen}]{qu2024}
Changle Qu, Sunhao Dai, Xiaochi Wei, Hengyi Cai, Shuaiqiang Wang, Dawei Yin,
  Jun Xu, and Ji-Rong Wen. 2024.
\newblock \href {https://arxiv.org/abs/2405.16089} {{Towards
  completeness-oriented tool retrieval for large language models}}.
\newblock In \emph{Proceedings of the 33rd ACM International Conference on
  Information and Knowledge Management}.

\bibitem[{Repantis et~al.(2026)Repantis, Singh, Joseph, Zhang, Vishwakarma,
  Karslioglu, Thot, and Gawde}]{repantis2026}
Vyzantinos Repantis, Harshvardhan Singh, Tony Joseph, Cien Zhang, Akash
  Vishwakarma, Svetlana Karslioglu, Michael~Wyatt Thot, and Ameya Gawde. 2026.
\newblock \href {https://arxiv.org/abs/2605.18857} {{The 99\% success paradox:
  When near-perfect retrieval equals random selection}}.
\newblock In \emph{ICLR 2026 Blog Track}.

\bibitem[{Robertson and Zaragoza(2009)}]{robertson2009}
Stephen Robertson and Hugo Zaragoza. 2009.
\newblock {The probabilistic relevance framework: BM25 and beyond}.
\newblock Foundations and Trends in Information Retrieval, 3(4).

\bibitem[{Saha et~al.(2026)Saha, Wang, and Manoharan}]{saha2026}
Olympia Saha, Amy Wang, and Srinivasan Manoharan. 2026.
\newblock \href {https://arxiv.org/abs/2608.23992} {{Hybrid semantic tool
  discovery for enterprise MCP gateway: Architecture and implementation}}.
\newblock \emph{Preprint}, arXiv:2608.23992.

\bibitem[{Sakizli(2026)}]{sakizli2026}
Furkan Sakizli. 2026.
\newblock \href {https://arxiv.org/abs/2605.26165} {{Tool-schema compression
  enables agentic RAG under constrained context budgets}}.
\newblock \emph{Preprint}, arXiv:2605.26165.

\bibitem[{Schick et~al.(2023)Schick, Dwivedi-Yu, Dess{\`\i}, Raileanu, Lomeli,
  Hambro, Zettlemoyer, Cancedda, and Scialom}]{schick2023}
Timo Schick, Jane Dwivedi-Yu, Roberto Dess{\`\i}, Roberta Raileanu, Maria
  Lomeli, Eric Hambro, Luke Zettlemoyer, Nicola Cancedda, and Thomas Scialom.
  2023.
\newblock \href {https://arxiv.org/abs/2302.04761} {{Toolformer: Language
  models can teach themselves to use tools}}.
\newblock In \emph{Advances in Neural Information Processing Systems 36}.

\bibitem[{Shen et~al.(2024)Shen, Song, Tan, Zhang, Ren, Yuan, Lu, Li, and
  Zhuang}]{shen2024}
Yongliang Shen, Kaitao Song, Xu~Tan, Wenqi Zhang, Kan Ren, Siyu Yuan, Weiming
  Lu, Dongsheng Li, and Yueting Zhuang. 2024.
\newblock \href {https://arxiv.org/abs/2311.18760} {{TaskBench: Benchmarking
  large language models for task automation}}.
\newblock In \emph{Advances in Neural Information Processing Systems 37}.

\bibitem[{Shi et~al.(2025)Shi, Wang, Yan, Ren, Wang, Yin, and Ren}]{shi2025}
Zhengliang Shi, Yuhan Wang, Lingyong Yan, Pengjie Ren, Shuaiqiang Wang, Dawei
  Yin, and Zhaochun Ren. 2025.
\newblock \href {https://arxiv.org/abs/2503.01763} {{Retrieval models aren't
  tool-savvy: Benchmarking tool retrieval for large language models}}.
\newblock In \emph{Findings of the Association for Computational Linguistics:
  ACL 2025}.

\bibitem[{Wang et~al.(2025)Wang, Han, Ji, Wang, Baldwin, and Li}]{wang2025}
Renxi Wang, Xudong Han, Lei Ji, Shu Wang, Timothy Baldwin, and Haonan Li. 2025.
\newblock \href {https://arxiv.org/abs/2410.03439} {{ToolGen: Unified tool
  retrieval and calling via generation}}.
\newblock In \emph{The Thirteenth International Conference on Learning
  Representations}.

\bibitem[{Weinberger and Hozez(2026)}]{weinberger2026}
Sarel Weinberger and Amir Hozez. 2026.
\newblock \href {https://arxiv.org/abs/2607.12161} {{Token reduction is not
  cost reduction: An empirical study of end-to-end efficiency in API-based
  coding agents}}.
\newblock \emph{Preprint}, arXiv:2607.12161.

\bibitem[{Wu et~al.(2024)Wu, Zhu, Han, Tan, Zhang, and Chen}]{wu2024}
Mengsong Wu, Tong Zhu, Han Han, Chuanyuan Tan, Xiang Zhang, and Wenliang Chen.
  2024.
\newblock \href {https://arxiv.org/abs/2405.08355} {{Seal-Tools: Self-instruct
  tool learning dataset for agent tuning and detailed benchmark}}.
\newblock \emph{Preprint}, arXiv:2405.08355.

\bibitem[{Xiao et~al.(2023)Xiao, Liu, Zhang, and Muennighoff}]{xiao2023}
Shitao Xiao, Zheng Liu, Peitian Zhang, and Niklas Muennighoff. 2023.
\newblock \href {https://arxiv.org/abs/2309.07597} {{C-Pack: Packed Resources
  For General Chinese Embeddings}}.
\newblock \emph{Preprint}, arXiv:2309.07597.

\bibitem[{Zhang et~al.(2024)Zhang, Fan, Wang, Chen, Mo, Sakai, and
  Yamana}]{zhang2024}
Yuxiang Zhang, Xin Fan, Junjie Wang, Chongxian Chen, Fan Mo, Tetsuya Sakai, and
  Hayato Yamana. 2024.
\newblock \href {https://arxiv.org/abs/2410.03212} {{Data-efficient massive
  tool retrieval: A reinforcement learning approach for query-tool alignment
  with language models}}.
\newblock \emph{Preprint}, arXiv:2410.03212.

\bibitem[{Zheng et~al.(2024)Zheng, Li, Liu, Liu, Luan, and Wang}]{zheng2024}
Yuanhang Zheng, Peng Li, Wei Liu, Yang Liu, Jian Luan, and Bin Wang. 2024.
\newblock \href {https://arxiv.org/abs/2403.06551} {{ToolRerank: Adaptive and
  hierarchy-aware reranking for tool retrieval}}.
\newblock In \emph{Proceedings of the 2024 Joint International Conference on
  Computational Linguistics, Language Resources and Evaluation (LREC-COLING
  2024)}.

\end{thebibliography}

\appendix

\section{Per-Run Results}\label{app:A}

\paragraph{Per-run results.}
Table~\ref{tab:4} lists every run of \S\ref{sec:4.2}. The eleven Codex runs used four variants: instructions to the agent on both gateways (three runs); the evaluation's original configuration, with those instructions on the recommendation gateway only (three runs); the original configuration with the order of the two conditions alternated (two runs); and the original configuration with one added sentence asking the agent to search for one action per task (three runs). Every run's total bill was lower with recommendation (sign test across runs, p = 0.0001).

\begin{table*}[t]
\centering
\small
\setlength{\tabcolsep}{4pt}
\begin{adjustbox}{max width=\textwidth}
\begin{tabular}{lrrlr}
\toprule
Client and variant & Run & Pairs passing both & Input, median change [95\% CI] & Total bill, all pairs \\
\midrule
Codex, instructions on both gateways & 1 & 24 & \ensuremath{-}12.1\% [\ensuremath{-}20.8, \ensuremath{-}1.8] & \ensuremath{-}4.3\% \\
Codex, instructions on both gateways & 2 & 25 & \ensuremath{-}27.2\% [\ensuremath{-}46.4, \ensuremath{-}11.4] & \ensuremath{-}4.9\% \\
Codex, instructions on both gateways & 3 & 26 & \ensuremath{-}28.0\% [\ensuremath{-}36.5, \ensuremath{-}1.4] & \ensuremath{-}6.8\% \\
Codex, original configuration & 1 & 19 & \ensuremath{-}6.5\% [\ensuremath{-}31.7, +11.3] & \ensuremath{-}2.9\% \\
Codex, original configuration & 2 & 22 & \ensuremath{-}0.1\% [\ensuremath{-}26.1, +20.7] & \ensuremath{-}6.5\% \\
Codex, original configuration & 3 & 19 & \ensuremath{-}25.7\% [\ensuremath{-}37.6, +9.9] & \ensuremath{-}7.7\% \\
Codex, alternating condition order & 1 & 21 & \ensuremath{-}26.3\% [\ensuremath{-}38.5, \ensuremath{-}14.0] & \ensuremath{-}3.7\% \\
Codex, alternating condition order & 2 & 21 & \ensuremath{-}15.5\% [\ensuremath{-}27.0, +0.8] & \ensuremath{-}5.9\% \\
Codex, one action per search task & 1 & 20 & \ensuremath{-}10.6\% [\ensuremath{-}36.8, +14.7] & \ensuremath{-}5.8\% \\
Codex, one action per search task & 2 & 21 & \ensuremath{-}14.3\% [\ensuremath{-}32.8, +9.1] & \ensuremath{-}2.6\% \\
Codex, one action per search task & 3 & 24 & \ensuremath{-}21.9\% [\ensuremath{-}33.1, \ensuremath{-}12.1] & \ensuremath{-}9.2\% \\
Claude Code, 21 suite workflows & 1 & 38 & \ensuremath{-}28.7\% [\ensuremath{-}31.6, \ensuremath{-}10.1] & \ensuremath{-}7.9\% \\
Claude Code, large tool results & 1 & 21 & \ensuremath{-}74.7\% [\ensuremath{-}77.3, \ensuremath{-}69.2] & \ensuremath{-}40.0\% \\
Claude Code, mixed result sizes & 1 & 15 & \ensuremath{-}48.0\% [\ensuremath{-}54.3, \ensuremath{-}31.5] & \ensuremath{-}14.0\% \\
\bottomrule
\end{tabular}
\end{adjustbox}
\setcounter{table}{3}
\caption{Per-run session input and bill. Except in the runs marked as having instructions on both gateways, only the recommendation gateway carried the instructions to the agent.}
\label{tab:4}
\end{table*}

\section{Benchmark Notes}\label{app:B}

The released ToolRet data differs from the paper, which describes 7,615 queries over 43,215 tools, and averages 1.77 relevant tools per query against 2.17. Our harness scores bge-base-en-v1.5 \citep{xiao2023}, without instructions over the full index as in \citet[Table~4]{shi2025}, at 27.3 nDCG@10 on the category average against 22.1 published, mostly on web (32.1 against 22.5).

\begin{table*}[t]
\centering
\small
\setlength{\tabcolsep}{4pt}
\begin{adjustbox}{max width=\textwidth}
\begin{tabular}{lrrrrrrrrr}
\toprule
 & \multicolumn{2}{c}{Web} & \multicolumn{2}{c}{Code} & \multicolumn{2}{c}{Customized} & \multicolumn{2}{c}{Average} & \multicolumn{1}{c}{Overall} \\
\cmidrule(lr){2-3} \cmidrule(lr){4-5} \cmidrule(lr){6-7} \cmidrule(lr){8-9} \cmidrule(lr){10-10}
System & R@10 & C@10 & R@10 & C@10 & R@10 & C@10 & R@10 & C@10 & C@10 \\
\midrule
\multicolumn{10}{l}{\emph{Arcade harness}} \\
Tool Recommendation & \textbf{56.81} & \textbf{45.60} & \textbf{57.25} & \textbf{54.26} & \textbf{42.65} & \textbf{27.19} & \textbf{52.24} & \textbf{42.35} & \textbf{45.23} \\
\ensuremath{\Delta} vs BM25 & +40.7\% & +47.5\% & +74.6\% & +79.4\% & +60.7\% & +80.4\% & +57.2\% & +66.7\% & +57.0\% \\
BM25 & 40.39 & 30.92 & 32.78 & 30.25 & 26.54 & 15.07 & 33.24 & 25.41 & 28.82 \\
\midrule
\multicolumn{10}{l}{\emph{ToolRet harness, without instructions\textsuperscript{\dag}}} \\
bge-reranker-v2-gemma & \textbf{45.94} & \textbf{27.85} & 38.23 & 36.82 & \textbf{49.43} & 37.75 & \textbf{44.53} & \textbf{34.14} & -- \\
\ensuremath{\Delta} vs BM25s & +86.6\% & +83.2\% & +35.4\% & +36.6\% & +52.6\% & +54.7\% & +56.7\% & +53.9\% &  \\
NV-Embed-v1 & 39.15 & 23.05 & \textbf{40.45} & \textbf{38.88} & 45.93 & 34.44 & 41.84 & 32.12 & -- \\
\ensuremath{\Delta} vs BM25s & +59.0\% & +51.6\% & +43.3\% & +44.2\% & +41.8\% & +41.1\% & +47.3\% & +44.8\% &  \\
GritLM-7B & 34.27 & 21.28 & 30.41 & 28.87 & 49.34 & \textbf{38.17} & 38.01 & 29.44 & -- \\
\ensuremath{\Delta} vs BM25s & +39.2\% & +40.0\% & +7.7\% & +7.1\% & +52.3\% & +56.4\% & +33.8\% & +32.7\% &  \\
BM25s & 24.62 & 15.20 & 28.23 & 26.96 & 32.39 & 24.40 & 28.41 & 22.19 & -- \\
\midrule
\multicolumn{10}{l}{\emph{Reported by Microsoft\textsuperscript{\dag}}} \\
Foundry tool search & 45.99 & -- & 39.56 & -- & 41.36 & -- & 42.30 & -- & -- \\
\ensuremath{\Delta} vs BM25s & +86.8\% &  & +40.1\% &  & +27.7\% &  & +48.9\% &  &  \\
\bottomrule
\end{tabular}
\end{adjustbox}
\setcounter{table}{4}
\caption{ToolRet results by category: Recall@10 (R) and Completeness@10 (C). Average is the mean of the three categories; Overall weights each query equally. Bold marks the best result in each block; scores from different harnesses are not directly comparable (\S\ref{sec:3.1}). Published rows use ToolRet's setting without instructions \citep[Table~4]{shi2025}. \textsuperscript{\dag}Quoted, not reproduced.}
\label{tab:5}
\end{table*}

\end{document}